\documentclass[10pt,twocolumn,letterpaper]{article}

\usepackage[pagenumbers]{wacv} 

\usepackage{graphicx}
\usepackage{amsmath}
\usepackage{amssymb}
\usepackage{booktabs}
\usepackage{array}
\usepackage{tikz}
\usepackage[inkscapelatex=false]{svg}
\usepackage{multirow}
\usepackage[ruled,vlined]{algorithm2e}
\usepackage{pifont} 

\usepackage[pagebackref,breaklinks,colorlinks]{hyperref}

\usepackage[%
  xindy,
  acronym,
]{glossaries}
\glsdisablehyper

\usepackage{siunitx}

\DeclareSIUnit\px{px}

\usepackage[capitalize]{cleveref}
\crefname{section}{Sec.}{Secs.}
\Crefname{section}{Section}{Sections}
\Crefname{table}{Table}{Tables}
\crefname{table}{Tab.}{Tabs.}

\usepackage[capitalize]{cleveref}
\crefname{section}{Sec.}{Secs.}
\Crefname{section}{Section}{Sections}
\Crefname{table}{Table}{Tables}
\crefname{table}{Tab.}{Tabs.}

\newacronym{gl:DEM}{DEM}{digital elevation model}
\newacronym{gl:ICAM}{ICAM}{image context attention module}
\newacronym{gl:TFG-RPAB}{TFG-RPAB}{terrain feature-guided residual pixel attention block}
\newacronym{gl:RMSE}{RMSE}{root mean square error} 
\newacronym{gl:DSM}{DSM}{Digital Surface Model}
\newacronym{gl:DTM}{DTM}{Digital Terrain Model}
\newacronym{gl:IDW}{IDW}{inverse distance weighting}
\newacronym{gl:GAN}{GAN}{Generative Adversarial Network}
\newacronym{gl:LiDAR}{LiDAR}{Light Detection and Ranging} 
\newacronym{gl:SRTM}{SRTM}{Shuttle Radar Topography Mission}
\newacronym{gl:FF}{FF}{Fill and Feather}
\newacronym{gl:DSF}{DSF}{Delta Surface Fill}
\newacronym{gl:GSD}{GSD}{Ground Sampling Distance}
\newacronym{gl:NMAD}{NMAD}{normalized median absolute deviation} 
\newacronym{gl:MedAE}{MedAE}{median absolute error}
\newacronym{gl:DDPM}{DDPM}{Denoising Diffusion Probabilistic Model}

\def\wacvPaperID{*****} 
\def\confName{WACV}
\def\confYear{2025}

\begin{document}

\title{Dfilled: Repurposing Edge-Enhancing  Diffusion for Guided DSM Void Filling}

\author{
    \parbox{\textwidth}{\centering
        Daniel Panangian\thanks{Corresponding author}\hspace{1cm}
        Ksenia Bittner \\
        The Remote Sensing Technology Institute \\  
        German Aerospace Center (DLR), Wessling, Germany \\
        \tt\small{\{daniel.panangian, ksenia.bittner\}@dlr.de}
    }
}
\maketitle

\begin{abstract}
\Glspl{gl:DSM} are essential for accurately representing Earth's topography in geospatial analyses. \Glspl{gl:DSM} capture detailed elevations of natural and man-made features, crucial for applications like urban planning, vegetation studies, and 3D reconstruction. However, \glspl{gl:DSM} derived from stereo satellite imagery often contain voids or missing data due to occlusions, shadows, and low-signal areas.  Previous studies have primarily focused on void filling for \glspl{gl:DEM} and \glspl{gl:DTM}, employing methods such as \gls{gl:IDW}, kriging, and spline interpolation. While effective for simpler terrains, these approaches often fail to handle the intricate structures present in \glspl{gl:DSM}. To overcome these limitations, we introduce  \textsc{Dfilled}, a guided \gls{gl:DSM} void filling method that leverages optical remote sensing images through edge-enhancing diffusion. Dfilled repurposes deep anisotropic diffusion models, which originally designed for super-resolution tasks, to inpaint \glspl{gl:DSM}. Additionally, we utilize Perlin noise to create inpainting masks that mimic natural void patterns in \glspl{gl:DSM}.
Experimental evaluations demonstrate that Dfilled surpasses traditional interpolation methods and deep learning approaches in \gls{gl:DSM} void filling tasks. Both quantitative and qualitative assessments highlight the method's ability to manage complex features and deliver accurate, visually coherent results.
\end{abstract}

\glsresetall
\section{Introduction}\label{MANUSCRIPT}

\begin{figure}[t]
    \centering
    \includegraphics[width=8.5cm]{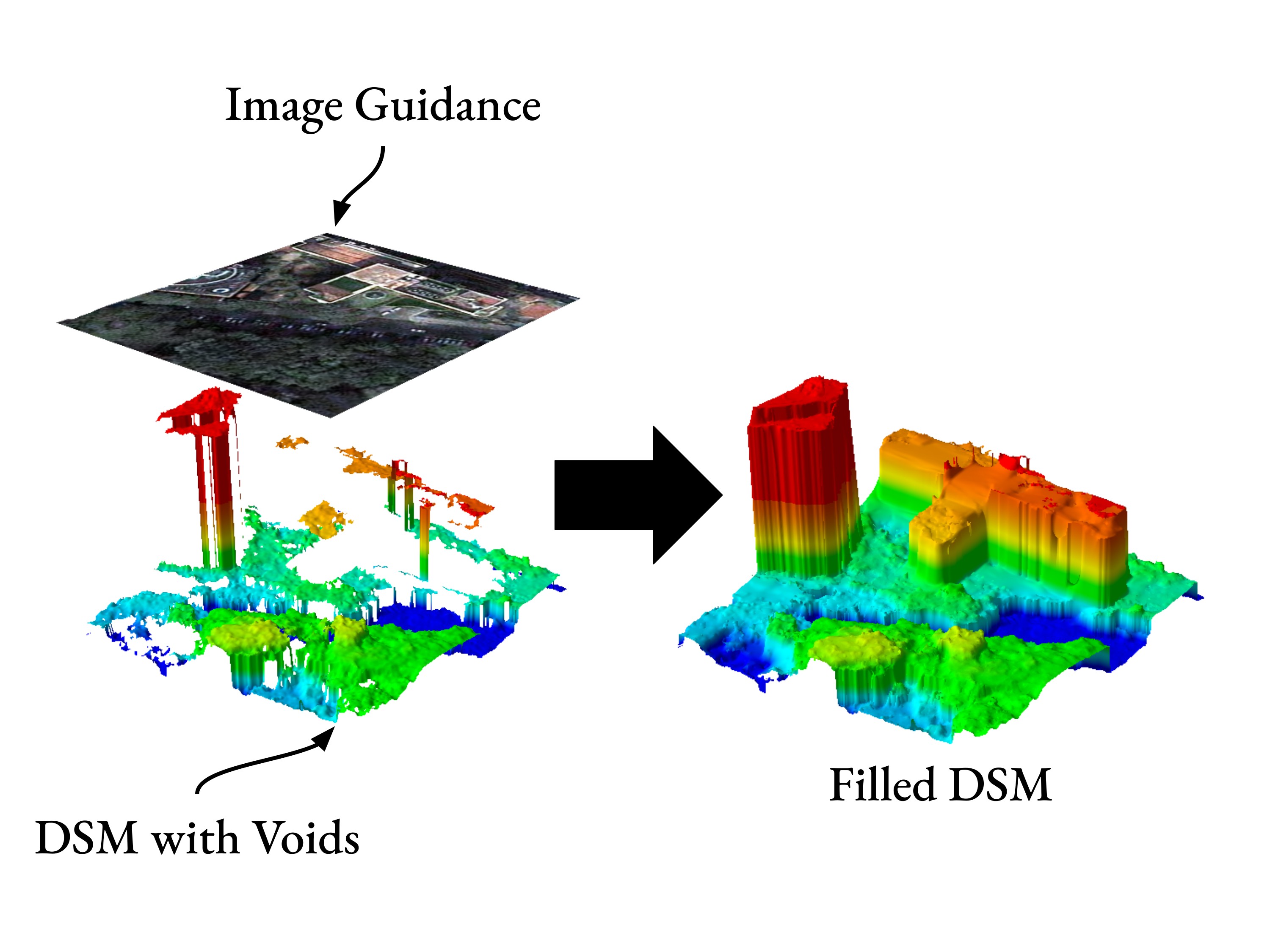} 
    \caption{Overview of the proposed guided DSM void filling approach. The method utilizes a image guidance to fill voids in the DSM, resulting in a complete and accurate DSM reconstruction.}
    \label{fig:overview}
\end{figure}

\Glspl{gl:DEM}—encompassing both \glspl{gl:DSM} and \glspl{gl:DTM}—are essential tools in geospatial analysis, representing the Earth's topography. \Glspl{gl:DSM}, in particular, include elevations of natural and man-made features such as vegetation and buildings, offering more detailed and complex information than \glspl{gl:DTM}. This richness makes \glspl{gl:DSM} invaluable for applications requiring fine-scale surface features, including urban planning, vegetation analysis, and 3D reconstruction.

However, \glspl{gl:DSM} generated from stereo satellite imagery often suffer from voids or holes (areas with missing elevation data) due to mismatches in occluded, shadowed, or low-signal regions. These voids degrade the quality and reliability of \glspl{gl:DSM}, impacting critical tasks in photogrammetry and remote sensing like change detection, object recognition, and 3D modeling. Filling these voids is particularly challenging in \glspl{gl:DSM} compared to \glspl{gl:DTM} because of the additional complexity introduced by surface features such as buildings and trees.

Traditional void filling methods involve interpolation techniques estimate unknown values using spatially neighboring information~\cite{Reuter2007}. While these methods may be sufficient for small gaps in \glspl{gl:DTM}, they often struggle with the complex features present in \glspl{gl:DSM}, especially when dealing with large voids or intricate urban landscapes. The reliability of these methods diminishes as the void size increases or the terrain becomes more complex, leading to inaccurate or unrealistic surface representations.

To overcome these limitations, researchers have explored the use of auxiliary data sources, such as integrating multiple \glspl{gl:DEM} or utilizing remote sensing imagery. Techniques like fill-and-feather, delta surface fill, and moving window erosion have been developed to fuse data from various sources~\cite{Dowding2004, Grohman2006}. However, these approaches may falter in areas with unreliable auxiliary data or fail to capture the detailed features inherent in \glspl{gl:DSM}.

The advent of deep learning has opened new avenues for \gls{gl:DEM} void filling. \Glspl{gl:GAN}, in particular, have shown promise due to their ability to generate realistic textures and patterns. Various \gls{gl:GAN}-based methods have been proposed, including conditional \glspl{gl:GAN}~\cite{Dong2017}, Wasserstein \glspl{gl:GAN} with contextual attention mechanisms~\cite{Gavriil2018}, and multi-attention \glspl{gl:GAN}~\cite{Zhou2019}. Some approaches have incorporated topographic features like slope and relief degree to enhance the training process~\cite{Qiu2020}. An alternative strategy is to utilize optical remote sensing images as auxiliary data for void filling. Remote sensing imagery provides rich spectral and textural information that correlates with the surface features present in \glspl{gl:DSM}. For example, buildings, trees, and other structures visible in optical images correspond to features in \glspl{gl:DSM}, offering valuable information for reconstructing missing elevation data. Previous work has attempted to incorporate such imagery by using shadow maps or other features to guide \gls{gl:DEM} reconstruction~\cite{Dong2020}. Nonetheless, these methods often focus on \glspl{gl:DTM} or small missing areas and may not fully leverage the rich feature information inherent in remote sensing imagery. They may also lack the ability to preserve the fine-scale details of man-made structures and vegetation, which are critical in \gls{gl:DSM} applications.

By utilizing edge-enhancing diffusion techniques, our proposed method enhances edges with optical image's guidance (see \cref{fig:overview}), which is critical for maintaining structural integrity in \glspl{gl:DSM}.

The key contributions of this work are:
\begin{enumerate}
    \item We introduce a novel approach that adapts deep anisotropic diffusion models, which originally designed for super-resolution tasks, to the problem of \glspl{gl:DSM} void filling. By redefining the problem formulation using the heat equation and modifying the model to handle localized missing data, including local refinement strategies for coarse reconstruction, we effectively propagate contextual information into voids while preserving critical structural details.

    \item We employ Perlin noise to generate inpainting masks that simulate the natural void patterns found in \glspl{gl:DSM}. This ensures that the model is trained on realistic missing data scenarios and enhances its ability to generalize to real \gls{gl:DSM} voids.

    \item  We demonstrate the effectiveness of the proposed method through extensive experiments on various simulated and real \gls{gl:DSM} datasets. We propose the use of Perlin noise also for realistic evaluation. Our approach outperforms traditional interpolation techniques and state-of-the-art deep learning methods, particularly in handling complex features and providing accurate, visually realistic results in handling both small and large void filling in \gls{gl:DSM}.
\end{enumerate}

\section{Related Work}\label{sec:TITLE AND ABSTRACT BLOCK}

\begin{figure*}[ht]
\begin{center}
\includegraphics[width=17cm]{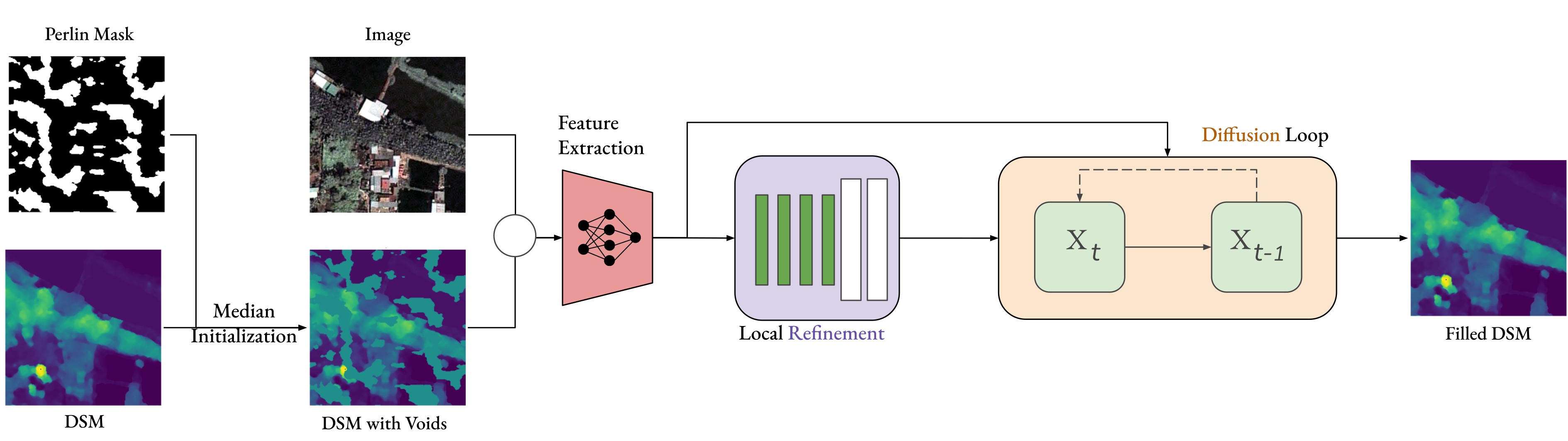}
\put(-327.5,61){$||$}
\end{center}
   \caption{Summary of the proposed architecture for DSM void filling. The method comprises a two-step process: First, high-dimensional features are extracted from both the void-filled DSM and high-resolution optical imagery using a pre-trained feature extractor. Next, a refinement network integrates residual blocks and upsampling operations to reconstruct missing elevation values. This is followed by an edge-enhancing diffusion network that iteratively refines the DSM, leveraging edge features from the optical imagery to ensure accurate and realistic reconstruction of terrain and structural details.}
\label{fig:realgdsr}
\end{figure*}

Voids in \glspl{gl:DEM} typically arise from limitations in data acquisition technologies such as radar or \gls{gl:LiDAR}. Factors like water bodies, dense vegetation, low reflectivity surfaces, and complex terrain can impede the collection of accurate elevation data. For example, radar-based missions like the \gls{gl:SRTM} are prone to issues like shadowing and layover in steep or rugged terrains, leading to data gaps \cite{Grohman2006}. Additionally, atmospheric conditions and instrument limitations can contribute to missing data in \glspl{gl:DEM}. In contrast, voids in \glspl{gl:DSM} often result from challenges inherent in stereo image matching processes used to generate \glspl{gl:DSM} from optical imagery. Occlusions caused by tall structures, shadows cast by buildings or terrain features, and regions with low texture or homogenous surfaces hinder the matching algorithms \cite{Krauss2012}. Urban environments with dense infrastructure and areas with significant vegetation present particular difficulties, leading to more frequent and extensive voids in \glspl{gl:DSM} compared to \glspl{gl:DTM}.

\subsection{Classical Methods for Void Filling}

To address the issue of voids in \glspl{gl:DEM} and \glspl{gl:DSM}, a variety of traditional methods have been developed. In \glspl{gl:DEM}, interpolation techniques such as \gls{gl:IDW}, kriging, and spline interpolation are commonly used \cite{Reuter2007}. These methods estimate missing elevation values based on the spatial correlation of surrounding data points. While effective for small voids in relatively flat and homogeneous terrains, their accuracy diminishes with increasing void size and terrain complexity. Many methods also have been proposed to leverage auxiliary data. The \gls{gl:FF}~\cite{Dowding2004} technique replaces missing data with values from an auxiliary DEM and applies smoothing at the edges to ensure seamless transitions. The \gls{gl:DSF}~\cite{Grohman2006} method creates a delta surface by computing the difference between the \gls{gl:DEM} and a resampled auxiliary surface, which is then used to adjust the fill surface for smooth integration. These methods rely heavily on the availability of high-quality auxiliary \glspl{gl:DEM}.

In the context of \glspl{gl:DSM}, traditional interpolation methods often struggle due to the complexity introduced by surface features like buildings and vegetation. To overcome this, Krau{\ss} \textit{et al.} \cite{Krauss2012} proposed a context-based approach for filling voids in \glspl{gl:DSM} generated from dense stereo matching. They categorized voids based on their characteristics and applied tailored strategies for each type, improving the accuracy of the filling process in complex urban areas. However, these methods may still be limited by the quality and availability of auxiliary data.

\subsection{Deep Learning-Based Methods}

The rise of deep learning has led to the development of more advanced void filling techniques for \glspl{gl:DEM}. \Glspl{gl:GAN} have been employed due to their ability to model complex data distributions and generate realistic outputs. Dong \textit{et al.}\cite{Dong2017} introduced a conditional \gls{gl:GAN} for recovering missing elevation data, demonstrating improved performance over traditional interpolation methods. Gavriil \textit{et al.}\cite{Gavriil2018} proposed a Wasserstein \gls{gl:GAN} with a contextual attention mechanism to enhance texture generation in void regions of \glspl{gl:DEM}. For \glspl{gl:DSM}, deep learning-based void filling methods are less prevalent but have shown potential. The complexity of \gls{gl:DSM} data, which includes detailed representations of surface features, poses significant challenges for modeling. Existing models often require extensive training data and may not fully exploit available information, such as high-resolution optical imagery, to improve void filling. Moreover, deep learning models may struggle to preserve fine-scale details crucial in \gls{gl:DSM} applications, such as edges of buildings and vegetation structures. Further enhancements have been achieved by integrating attention mechanisms into a conditional \gls{gl:GAN}, such as in the context-aware models proposed by Zhang \textit{et al.}\cite{Zhang2020} and Zhou \textit{et al.}\cite{Zhou2019}, which leverage multi-attention mechanisms to improve void filling performance. Additionally, domain-specific constraints have been introduced to guide restoration. For instance, Shadow-constrained GAN (SCGAN) incorporates terrain shadow geometry into its loss function \cite{ShadowGAN2021}, while the Topographic Knowledge-Constrained GAN (TKCGAN) penalizes incorrect valley and ridge predictions \cite{Zhou2021}. Recently, Lo \textit{et al.} \cite{Lo2024} proposed a novel approach using a conditional \gls{gl:DDPM} for void filling in \glspl{gl:DEM}. The Diff-DEM framework leverages the iterative refinement capabilities of \glspl{gl:DDPM} to generate high-quality reconstructions. This method represents a shift from traditional \glspl{gl:GAN} to probabilistic diffusion models, offering promising results in both accuracy and robustness.

\subsection{Use of Guidance in Void Filling Methods}

Incorporating guidance from auxiliary data modalities has proven effective in enhancing void filling techniques. For \glspl{gl:DEM}, remote sensing imagery provides valuable contextual information that aids in improving reconstruction quality. Dong \textit{et al.}\cite{Dong2020} utilized shadow maps extracted from optical images as constraints within a convolutional neural network, resulting in improved accuracy in void filling for \glspl{gl:DEM}. Similarly, Qiu \textit{et al.}\cite{Qiu2020} introduced a terrain texture generation model that integrates topographic features, such as slope and relief degree, into a \gls{gl:GAN} framework, enhancing the realism of generated terrain in void regions of \glspl{gl:DEM}. Building on these advancements, Yue \textit{et al.}\cite{Yue2024} proposed a terrain feature-guided transfer learning approach assisted by remote sensing images. This method leverages terrain features and auxiliary data to guide the generative process, achieving more accurate and realistic void filling in \glspl{gl:DEM}. In the case of \glspl{gl:DSM}, the use of auxiliary guidance remains a question. Optical imagery, rich in spectral and textural information, could serve as a valuable data source to capture surface feature details and improve reconstruction quality in \gls{gl:DSM} void filling tasks.

\section{Methodology}

\subsection{Repurposing Deep Anisotropic Diffusion for Inpainting}

Deep anisotropic diffusion models have demonstrated significant success in super-resolution tasks by enhancing spatial resolution through the reconstruction of high-frequency details from low-resolution images\cite{Metzger_2023_CVPR}. These models utilize anisotropic diffusion processes to propagate information directionally, effectively enhancing edges and fine details while mitigating noise and artifacts. Recognizing their potential in handling structural complexities, we repurpose deep anisotropic diffusion models for the task of image inpainting.

In image inpainting, the objective is to fill in missing or corrupted regions within an image in a manner that is coherent with the surrounding content. By adapting deep anisotropic diffusion models, we aim to effectively propagate contextual information from known regions into voids, resulting in seamless and visually plausible inpainted images (see \cref{fig:realgdsr}).

\subsection{Problem Formulation}

The inpainting problem can be formulated using the heat equation, which models the diffusion of heat (or, analogously, information) over time. The classical isotropic heat equation is given by:

\begin{equation}
\frac{\partial u}{\partial t} = \nabla^2 u,
\end{equation}
where \( u(x, y, t) \) represents the image intensity at position \( (x, y) \) and time \( t \), and \( \nabla^2 \) is the Laplacian operator.

For image inpainting, we are interested in finding a steady-state solution (\( \frac{\partial u}{\partial t} = 0 \)) to the anisotropic diffusion equation:

\begin{equation}
\frac{\partial u}{\partial t} = \nabla \cdot (D(x, y) \nabla u),
\end{equation}
where \( D(x, y) \) is the diffusion tensor that controls the rate and direction of diffusion at each point in the image. The diffusion tensor is designed to encourage diffusion along the edges and inhibit it across edges, preserving important structural details.

Our goal is to reconstruct the missing regions \( \Omega_{\text{missing}} \) in the image \( u(x, y) \) such that the inpainted image satisfies the anisotropic diffusion equation within \( \Omega_{\text{missing}} \) and agrees with the known pixel values in the known regions \( \Omega_{\text{known}} \):

\begin{equation}
\begin{cases}
\frac{\partial u}{\partial t} = \nabla \cdot (D(x, y) \nabla u), & \text{for } (x, y) \in \Omega_{\text{missing}}, \\
u(x, y) = u_0(x, y), & \text{for } (x, y) \in \Omega_{\text{known}},
\end{cases}
\end{equation}
where \( u_0(x, y) \) denotes the known pixel values.

By formulating the problem in terms of the heat equation, we can employ the deep anisotropic diffusion model to iteratively solve for \( u(x, y) \), effectively diffusing information from known regions into the missing areas.

\subsection{Proposed Model}

To tailor the deep anisotropic diffusion model for inpainting, we introduce the following key extensions:

\subsubsection{Coarse Reconstruction with Local Refinement Network}

The initialization of missing regions significantly impacts the diffusion performance. Inadequate initialization can lead to poor convergence or unrealistic artifacts in the final output \cite{realgdsr}. We implement the local refinement network for coarse reconstruction of the missing regions. This local refinement provides an informed starting point for the diffusion process, enabling the model to focus on refining details rather than constructing basic structures from scratch. To achieve this we implement the same network used in \cite{realgdsr} by removing the residual connection.

\subsubsection{Perlin Noise for Mask Generation}

\begin{figure}
    \begin{subfigure}{\linewidth}
    \centering
    \end{subfigure} 
        \begin{subfigure}{\linewidth}
        \centering
        \begin{subfigure}{.32\linewidth}
            \includegraphics[width=\linewidth]{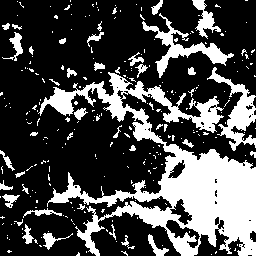}
        \end{subfigure}
        \begin{subfigure}{.32\linewidth}
            \includegraphics[width=\linewidth]{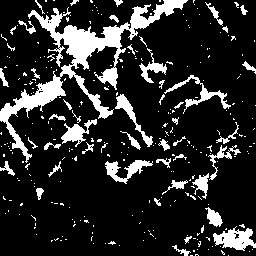}
        \end{subfigure}
        \begin{subfigure}{.32\linewidth}
            \includegraphics[width=\linewidth]{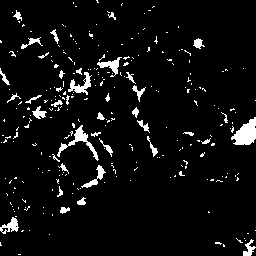}
        \end{subfigure}
        \caption{Real Voids}
    \end{subfigure} 
        \begin{subfigure}{\linewidth}
        \centering
        \begin{subfigure}{.32\linewidth}
            \includegraphics[width=\linewidth]{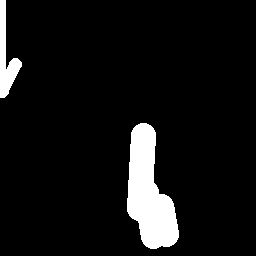}
        \end{subfigure}
        \begin{subfigure}{.32\linewidth}
            \includegraphics[width=\linewidth]{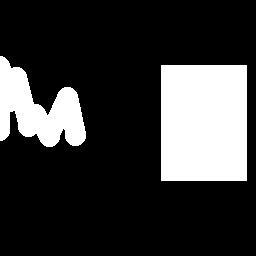}
        \end{subfigure}
        \begin{subfigure}{.32\linewidth}
            \includegraphics[width=\linewidth]{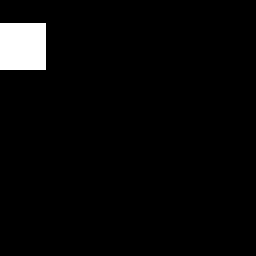}
        \end{subfigure}
        \caption{LaMa\cite{suvorov2021resolution} Mask}
    \end{subfigure} 
        \begin{subfigure}{\linewidth}
        \centering
        \begin{subfigure}{.32\linewidth}
            \includegraphics[width=\linewidth]{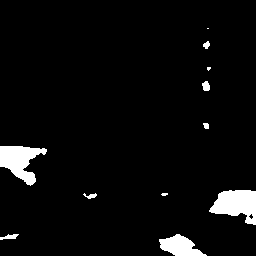}
        \end{subfigure}
        \begin{subfigure}{.32\linewidth}
            \includegraphics[width=\linewidth]{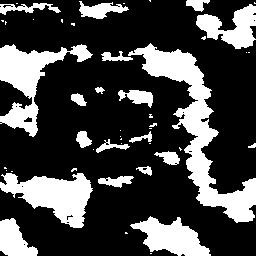}
        \end{subfigure}
        \begin{subfigure}{.32\linewidth}
            \includegraphics[width=\linewidth]{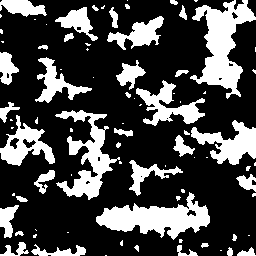}
        \end{subfigure}
        \caption{Perlin Mask }
    \end{subfigure} 
    \caption{Real and synthetic void masks for DSM void filling. (a) Real Voids illustrate naturally occurring, complex patterns in DSMs. (b) LaMa masks are structured synthetic voids often used in model training but may not accurately reflect real void distributions. (c) Perlin masks, generated with procedural noise, better mimic the irregularity and complexity of real voids.}
    \label{fig:real-low-dsm}
\end{figure}

An essential aspect of training and evaluating inpainting models is the design of masks that simulate missing regions \cite{suvorov2021resolution}. Previous methods often employ masks composed of irregular and rectangular shapes. While effective for general images, these masks may not accurately represent the void patterns found in \glspl{gl:DEM} and \glspl{gl:DSM}, which often exhibit natural, continuous missing regions due to occlusions or sensor limitations.

To address this, we utilize Perlin noise to generate masks that more closely resemble the voids in \glspl{gl:DEM} and \glspl{gl:DSM}, as seen in \cref{fig:real-low-dsm}. Perlin noise is a gradient noise function producing natural-looking textures with continuous gradients, widely used in procedural texture generation. Algorithm \ref{alg:perlinnoisemask} details the mask generation procedure.

\begin{algorithm}[t]
\caption{\textsc{PerlinMask}}
\label{alg:perlinnoisemask}
\DontPrintSemicolon

\KwIn{$s$: image size}
\KwOut{$\textit{mask}$}

Randomly select parameters: scale($sc$), octaves($o$), persistence($p$), lacunarity($l$), base($b$)\;
Initialize an $s \times s$ array, $\textit{noise}$\;
\For{$i=0$ \KwTo $s-1$}{
  \For{$j=0$ \KwTo $s-1$}{
    $\textit{noise}[i,j] \gets \text{PerlinNoise}(i/sc,\, j/sc,\, o,\, p,\, l,\, b)$
  }
}

Normalize $\textit{noise}$ to $[0,1]$\;
$\textit{threshold} \gets \text{Uniform}(0,1)$\;

Initialize $\textit{mask}$ as $s \times s$\;
\For{$i=0$ \KwTo $s-1$}{
  \For{$j=0$ \KwTo $s-1$}{
    $\textit{mask}[i,j] \gets \mathbf{1}[\textit{noise}[i,j] > \textit{threshold}]$
  }
}

\Return $\textit{mask}$

\end{algorithm}

\section{Experiments}

\newcolumntype{M}[1]{>{\centering\arraybackslash}m{#1}}
\begin{figure*}[!h]
    \centering
    \begin{tabular}{cM{50mm}M{50mm}M{50mm}}
        \toprule
                \toprule
        \textbf{} & \textbf{Area 1} & \textbf{Area 2} & \textbf{Area 3} \\
        \midrule
        \rotatebox[origin=c]{90}{\textbf{Input}} 
        & \begin{tikzpicture}
            \node[anchor=south west,inner sep=0] (image) at (0,0) {\includegraphics[height=4.7cm]{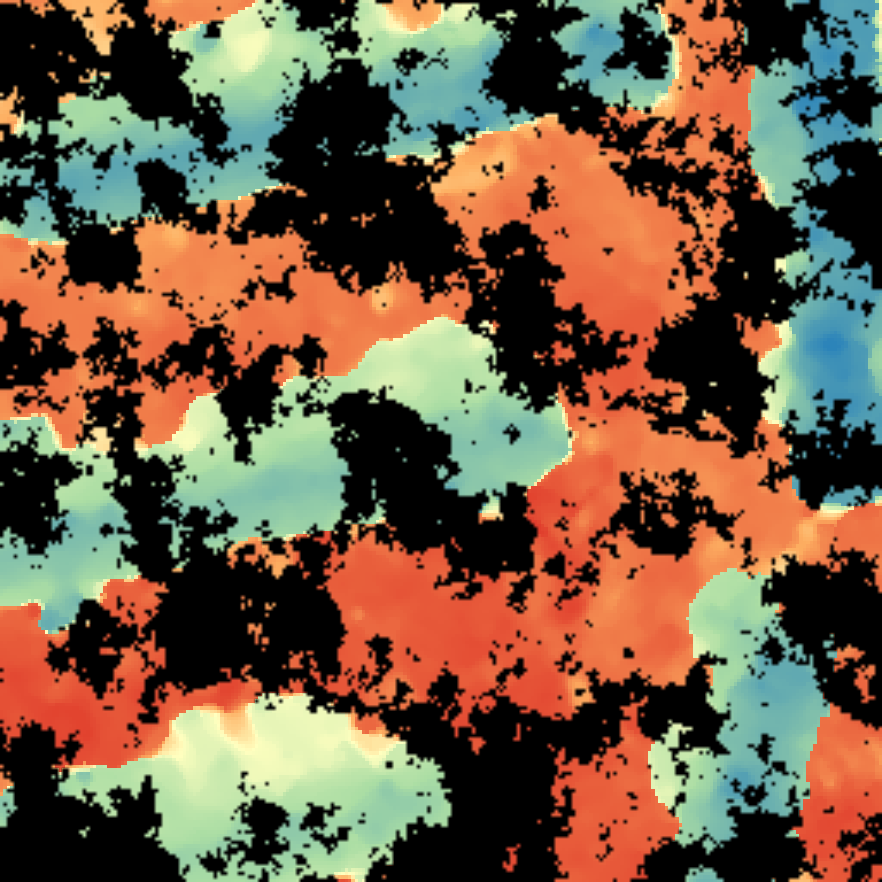}};
            \begin{scope}[x={(image.south east)},y={(image.north west)}]
                \draw[green, thick] (0.8,0.4) rectangle (0.9,0.9); 
                \draw[green, thick] (0.40,0.02) rectangle (0.6,0.2);
            \end{scope}
          \end{tikzpicture} 
        & \begin{tikzpicture}
            \node[anchor=south west,inner sep=0] (image) at (0,0) {\includegraphics[height=4.7cm]{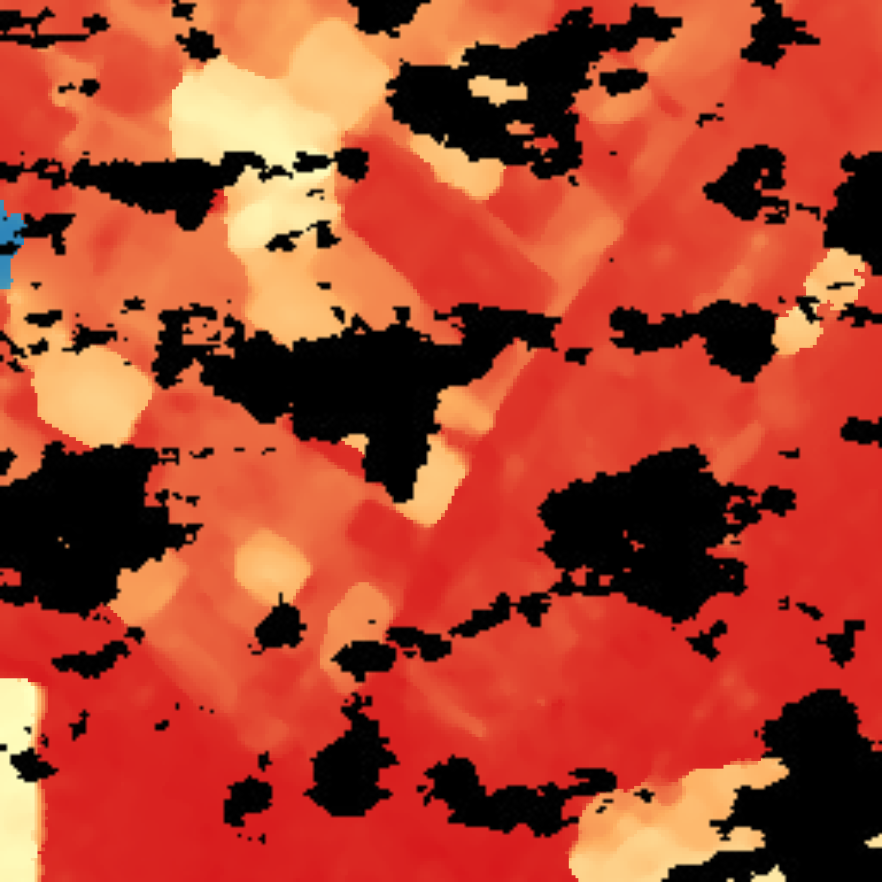}};
            \begin{scope}[x={(image.south east)},y={(image.north west)}]
                \draw[green, thick] (0.25,0.7) rectangle (0.5,0.5); 
            \end{scope}
          \end{tikzpicture} 
        & \begin{tikzpicture}
            \node[anchor=south west,inner sep=0] (image) at (0,0) {\includegraphics[height=4.7cm]{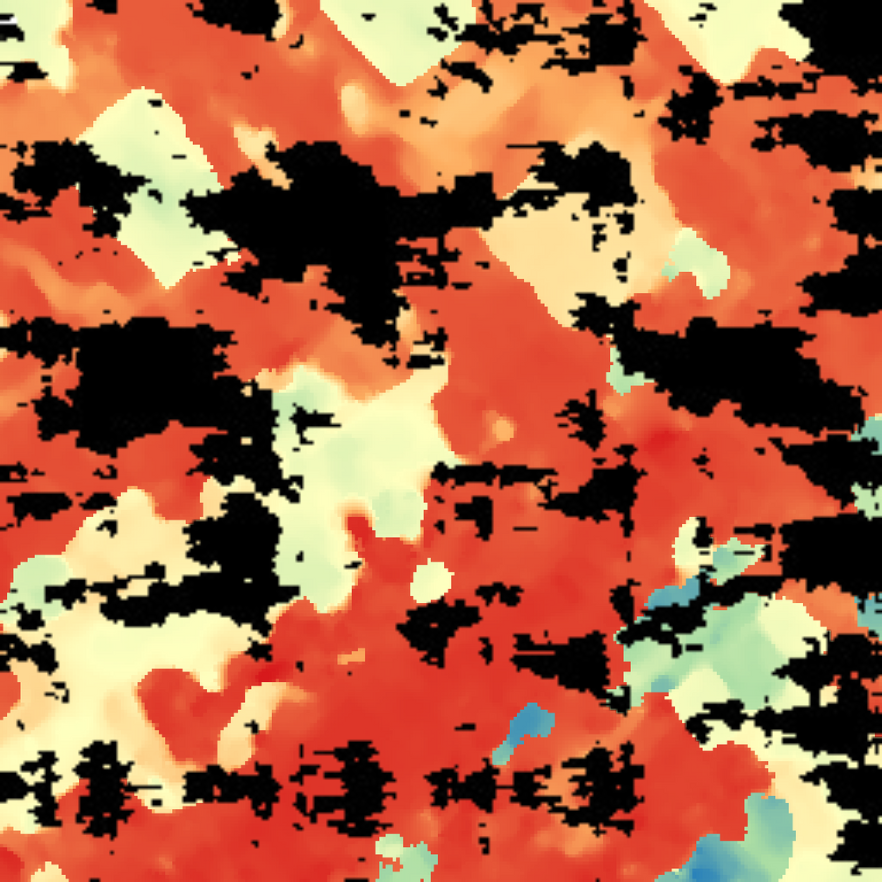}};
            \begin{scope}[x={(image.south east)},y={(image.north west)}]
                \draw[green, thick] (0.2,0.9) rectangle (0.45,0.65); 
            \end{scope}
          \end{tikzpicture} \\
        
        \rotatebox[origin=c]{90}{\textbf{RSAGAN}} 
        & \begin{tikzpicture}
            \node[anchor=south west,inner sep=0] (image) at (0,0) {\includegraphics[height=4.7cm]{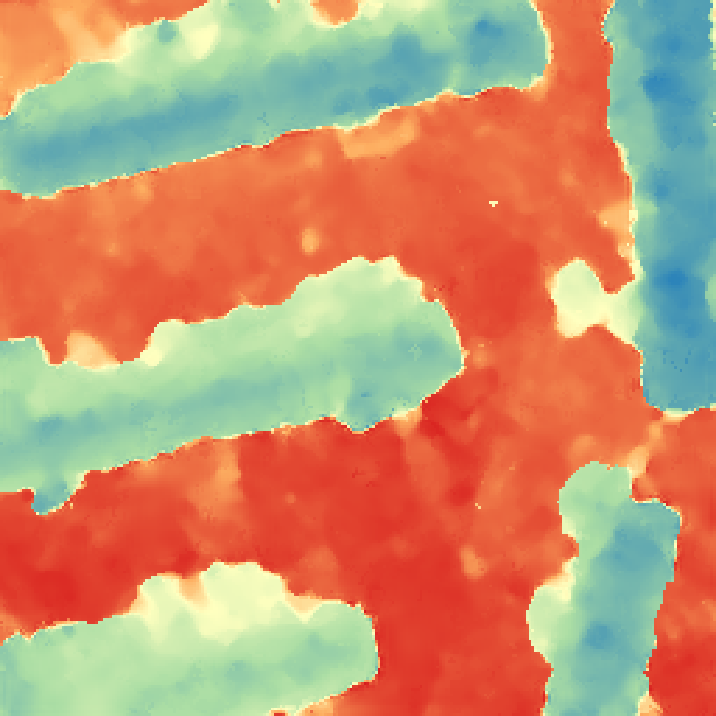}};
            \begin{scope}[x={(image.south east)},y={(image.north west)}]
                \draw[green, thick] (0.8,0.4) rectangle (0.9,0.9);  
                \draw[green, thick] (0.40,0.02) rectangle (0.6,0.2);
            \end{scope}
          \end{tikzpicture} 
        & \begin{tikzpicture}
            \node[anchor=south west,inner sep=0] (image) at (0,0) {\includegraphics[height=4.7cm]{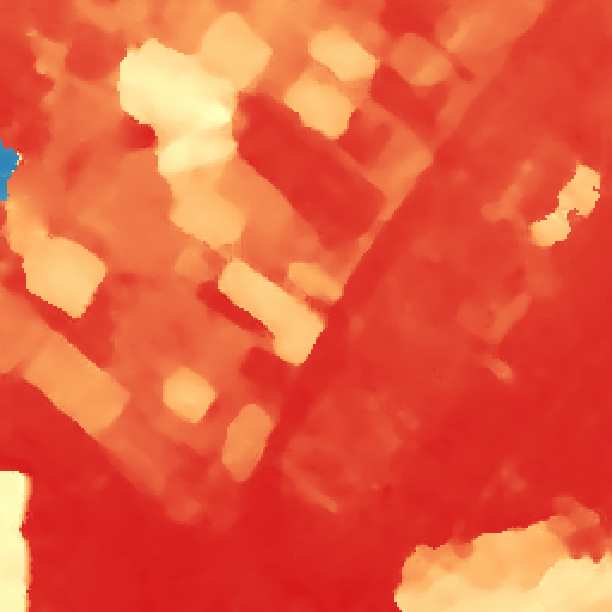}};
            \begin{scope}[x={(image.south east)},y={(image.north west)}]
                \draw[green, thick] (0.25,0.7) rectangle (0.5,0.5); 
            \end{scope}
          \end{tikzpicture} 
        & \begin{tikzpicture}
            \node[anchor=south west,inner sep=0] (image) at (0,0) {\includegraphics[height=4.7cm]{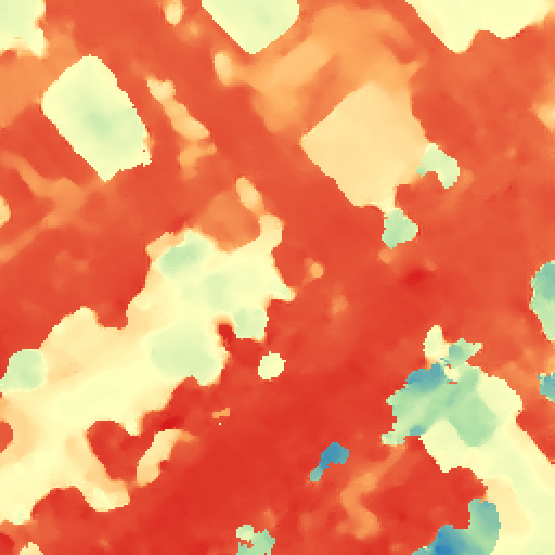}};
            \begin{scope}[x={(image.south east)},y={(image.north west)}]
                \draw[green, thick] (0.2,0.9) rectangle (0.45,0.65); 
            \end{scope}
          \end{tikzpicture} \\
        
        \rotatebox[origin=c]{90}{\textbf{Dfilled}} 
        & \begin{tikzpicture}
            \node[anchor=south west,inner sep=0] (image) at (0,0) {\includegraphics[height=4.7cm]{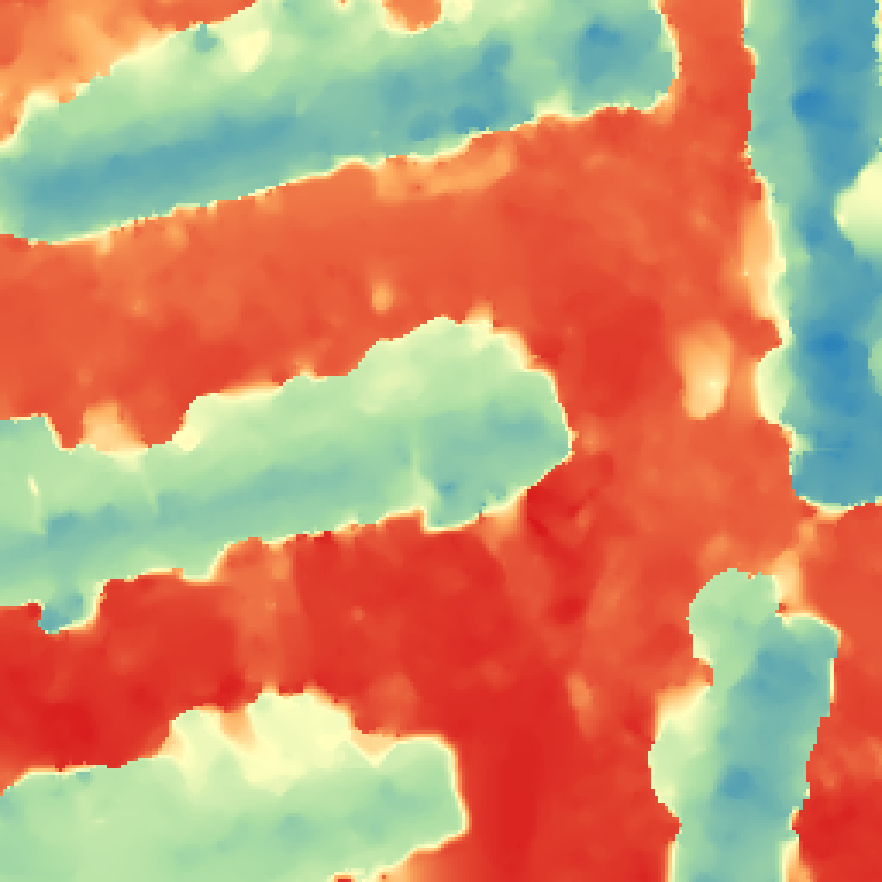}};
            \begin{scope}[x={(image.south east)},y={(image.north west)}]
                \draw[green, thick] (0.8,0.4) rectangle (0.9,0.9);  
                \draw[green, thick] (0.40,0.02) rectangle (0.6,0.2);
            \end{scope}
          \end{tikzpicture} 
        & \begin{tikzpicture}
            \node[anchor=south west,inner sep=0] (image) at (0,0) {\includegraphics[height=4.7cm]{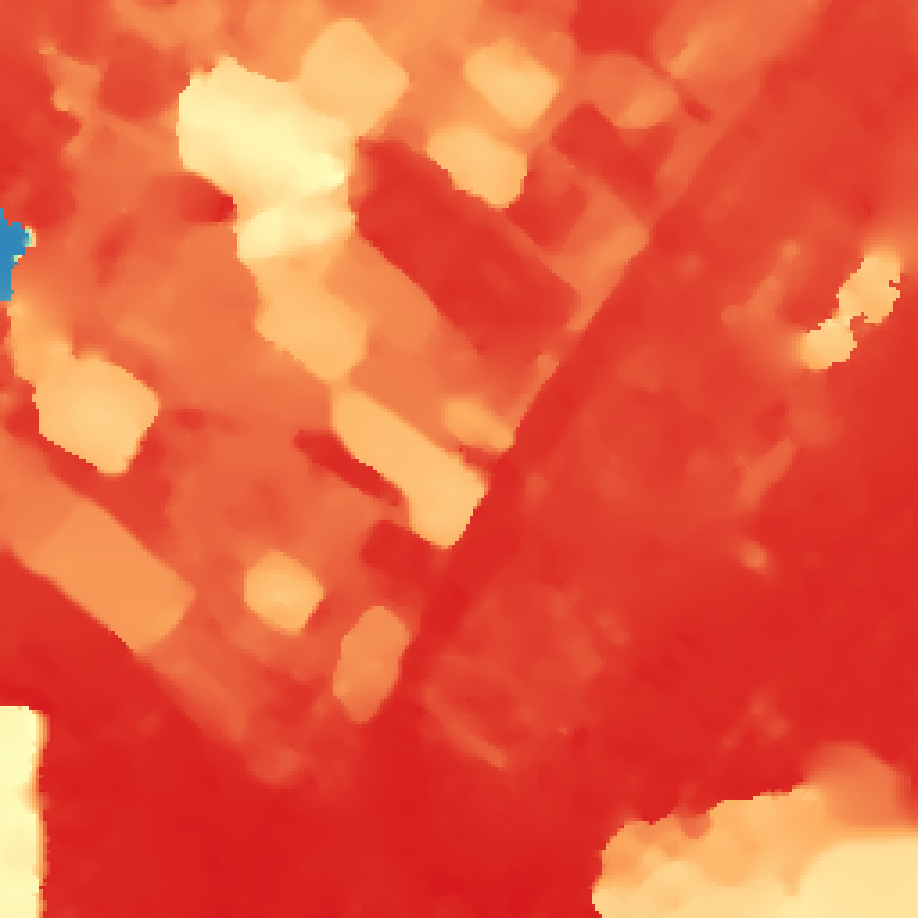}};
            \begin{scope}[x={(image.south east)},y={(image.north west)}]
                \draw[green, thick] (0.25,0.7) rectangle (0.5,0.5); 
            \end{scope}
          \end{tikzpicture} 
        & \begin{tikzpicture}
            \node[anchor=south west,inner sep=0] (image) at (0,0) {\includegraphics[height=4.7cm]{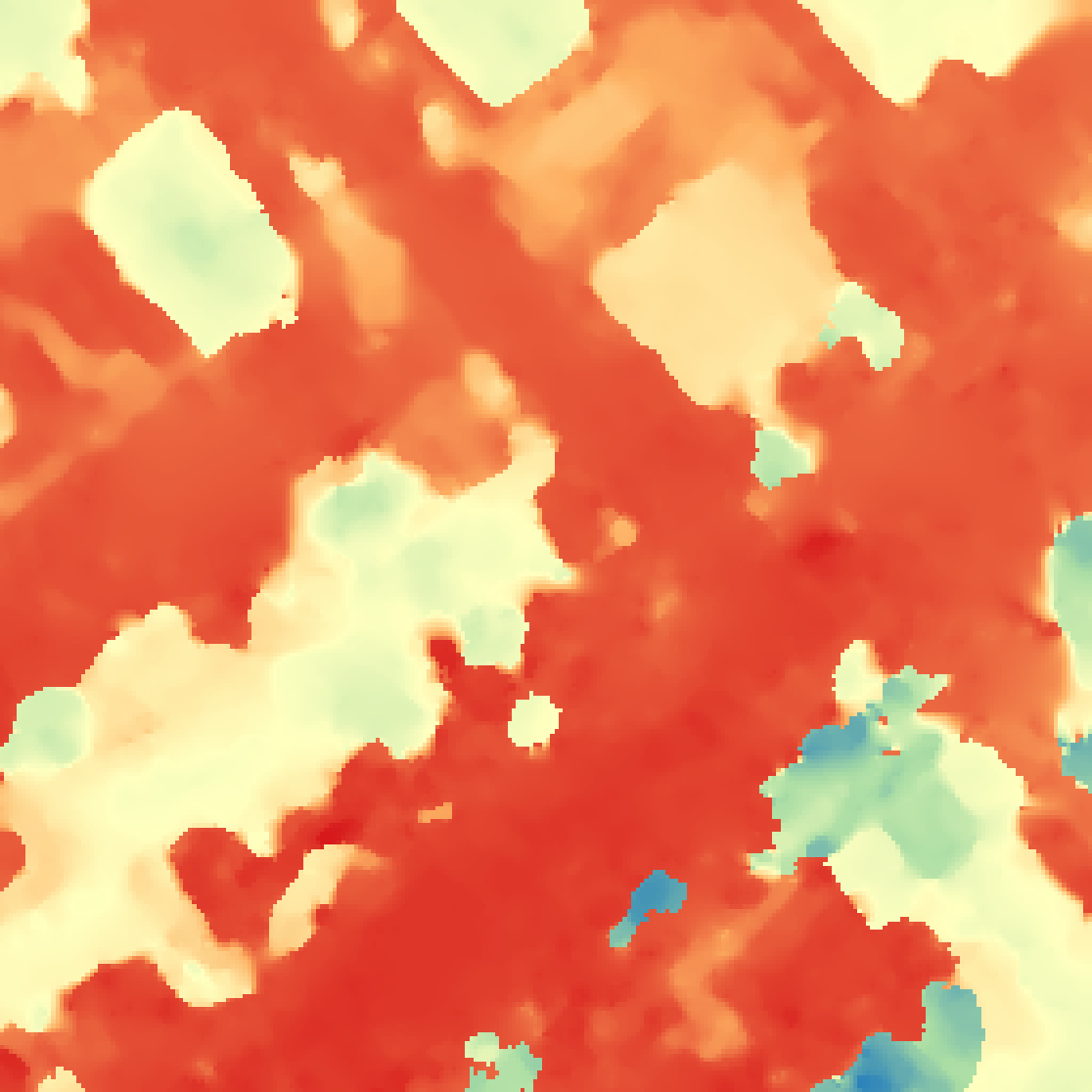}};
            \begin{scope}[x={(image.south east)},y={(image.north west)}]
               \draw[green, thick] (0.2,0.9) rectangle (0.45,0.65); 
            \end{scope}
          \end{tikzpicture} \\
        
        \rotatebox[origin=c]{90}{\textbf{Ground Truth}} 
        & \begin{tikzpicture}
            \node[anchor=south west,inner sep=0] (image) at (0,0) {\includegraphics[height=4.7cm]{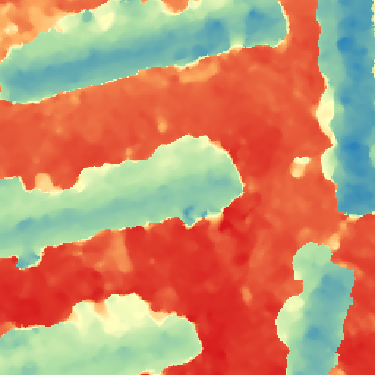}};
            \begin{scope}[x={(image.south east)},y={(image.north west)}]
                \draw[green, thick] (0.8,0.4) rectangle (0.9,0.9);  
                \draw[green, thick] (0.40,0.02) rectangle (0.6,0.2); 
            \end{scope}
          \end{tikzpicture} 
        & \begin{tikzpicture}
            \node[anchor=south west,inner sep=0] (image) at (0,0) {\includegraphics[height=4.7cm]{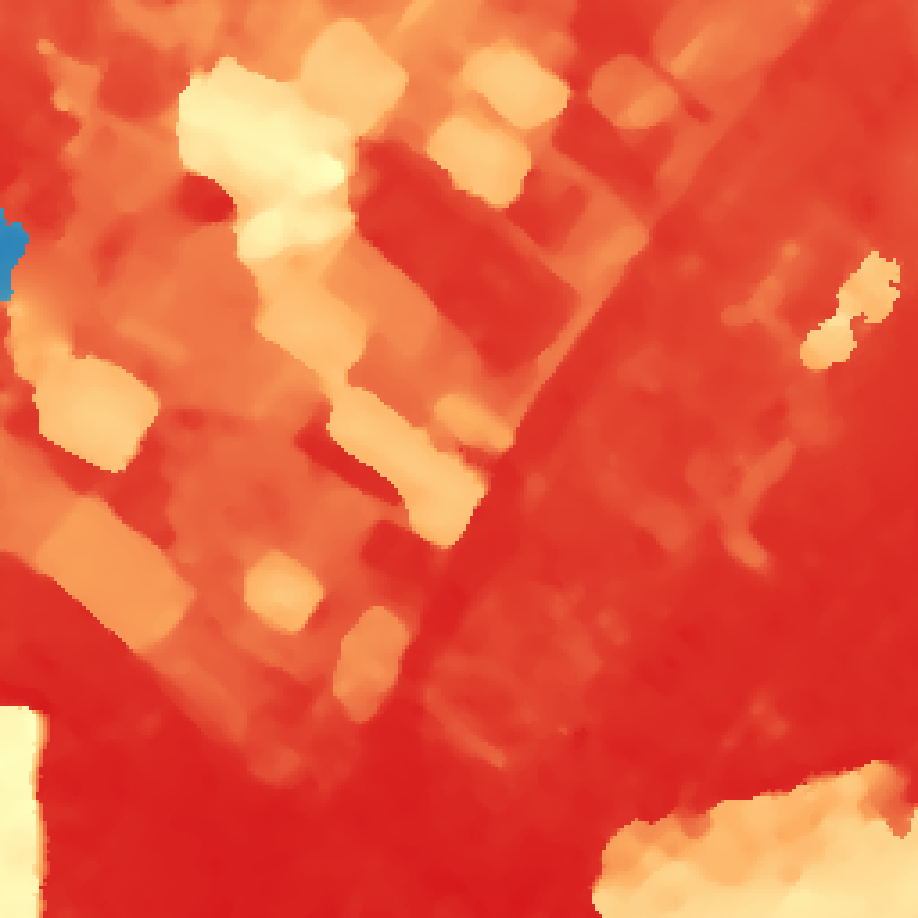}};
            \begin{scope}[x={(image.south east)},y={(image.north west)}]
                \draw[green, thick] (0.25,0.7) rectangle (0.5,0.5); 
            \end{scope}
          \end{tikzpicture} 
        & \begin{tikzpicture}
            \node[anchor=south west,inner sep=0] (image) at (0,0) {\includegraphics[height=4.7cm]{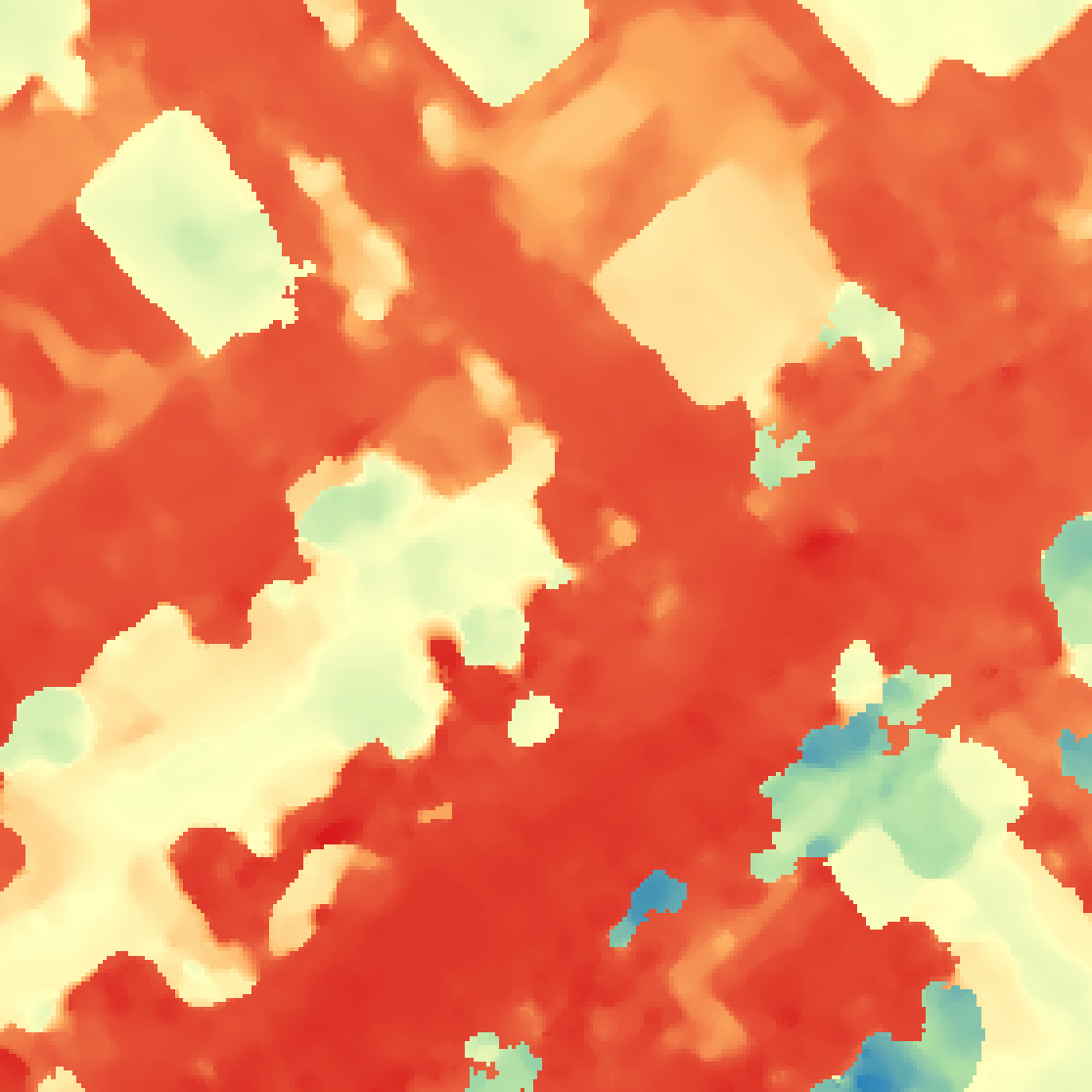}};
            \begin{scope}[x={(image.south east)},y={(image.north west)}]
                \draw[green, thick] (0.2,0.9) rectangle (0.45,0.65); 
            \end{scope}
          \end{tikzpicture} \\

        \bottomrule
    \end{tabular}
    \caption{Visual comparison of void filling results for small masks, where voids cover a relatively small portion of the DSM. The proposed Dfilled method produces more regularized and smoother results compared to RSAGAN, aligning closely with the ground truth while preserving fine-scale structural details. Green boxes indicate regions of interest.}
    \label{fig:small_masks}
\end{figure*}

\subsection{Datasets and Implementation}
\paragraph{Imagery and Study Area} 
We use \glspl{gl:DSM} and corresponding RGB orthoimages acquired over Ho Chi Minh City, Vietnam; Zurich and Bern, Switzerland; and Dushanbe, Tajikistan, all with a \gls{gl:GSD} of \SI{0.5}{\meter}. The \glspl{gl:DSM} for Ho Chi Minh City and Dushanbe are generated from Pleiades 1B satellite data using a single triplet stereo acquisition. The high-resolution DSMs for Switzerland are LiDAR DSMs provided by The Federal Office of Topography on the Swisstopo Portal\footnote{\url{https://www.swisstopo.admin.ch/en/geodata.html}} and are the same as those proposed in \cite{realgdsr}.

The dataset consists of approximately 4000 patches for training, 400 for validation, and 1300 for testing, each of size (\SI{256}{\px}, \SI{256}{\px}). We use Ho Chi Minh City and Switzerland data for training and validation, and use Dushanbe data for testing. The study areas encompass diverse urban environments, including widely spaced, detached residential buildings, allotments, and high commercial buildings.

\begin{table*}[!ht]
\caption{Comparison of models on real and synthetic datasets using RMSE, NMAD, and MedAE metrics. The table lists the model name, whether guide image guidance is incorporated and the training mask used. The performance metrics are evaluated on three different datasets: Real, Small Synthetic, and Large Synthetic.}
\centering
\scalebox{0.85}{
\begin{tabular}{l | c | l | c c c | c c c | c c c}
\toprule
\textbf{Model} & \textbf{Image} & \textbf{Train} & \multicolumn{3}{c|}{\textbf{Real}} & \multicolumn{3}{c|}{\textbf{Small Synthetic}} & \multicolumn{3}{c}{\textbf{Large Synthetic}} \\ 
\cmidrule(lr){4-6} \cmidrule(lr){7-9} \cmidrule(lr){10-12}
               & \textbf{Guidance}               & \textbf{Mask}                     & \textbf{RMSE} & \textbf{NMAD} & \textbf{MedAE} & \textbf{RMSE} & \textbf{NMAD} & \textbf{MedAE} & \textbf{RMSE} & \textbf{NMAD} & \textbf{MedAE} \\ 
\midrule
\textbf{Spline}      &     -           &    -    & 16.69       & 3.85         & 16.88       & 18.80       & 2.45         & 18.55       & 20.56       & 2.98         & 18.93       \\ 
\midrule
\textbf{Diff-DEM}    &     -           & LaMa    & 3.87        & 0.98         & 0.67        & 3.49        & 0.76         & 0.54        & 13.59       & 5.82         & 12.89       \\ 
                     &                & Perlin  & \textbf{3.02}        & 0.94         & 0.67        & 3.50        & 0.76         & 0.54        & 13.20       & 5.82         & 12.90       \\ 
\midrule
\textbf{RSAGAN}      & $\checkmark$   & LaMa    & 4.25        & 1.66         & 1.18        & 2.37        & 0.87         & 0.63        & 6.60        & 5.82         & 3.86        \\ 
                     &    & Perlin  & 3.71        & 0.83         & 0.57        & 1.28        & 0.27         & 0.19        & 3.62        & 1.57         & 1.27        \\ 
\midrule
\textbf{Dfilled (Ours)} & $\checkmark$ & LaMa    & \textbf{2.91}  & \textbf{0.30}  & \textbf{0.20}  & \textbf{1.14}  & \textbf{0.20}  & \textbf{0.14}  & \textbf{2.49}  & \textbf{0.78}  & \textbf{0.54}  \\ 
                       & & Perlin  & 3.17       & \textbf{0.49}         & \textbf{0.33}       & \textbf{1.23}       & \textbf{0.20}         & \textbf{0.14}       & \textbf{2.53}        & \textbf{0.97}         & \textbf{0.70}       \\
\bottomrule
\end{tabular}}
\label{tab:comparison}
\end{table*}

\paragraph{Implementation Details}
 We randomly load training patches during training. We initialize the voids in the \glspl{gl:DSM} with the median value of the known regions. We normalize the data using min-max normalization: for the refinement network, each \gls{gl:DSM} is normalized to the range [-1, 1]; for the diffusion network, \glspl{gl:DSM} are normalized to the range [0, 1]. Optical images are normalized using ImageNet statistics. In the training data, voids masks are introduced similarly to the method presented in \cite{liu2018image}, generating irregular mask shapes to simulate voids in the \glspl{gl:DSM}. The testing data includes real voids masks and corresponding ground truth.

In all experiments, we use a hidden feature dimension of 64 for the feature extractor and the refinement decoder. A ResNet-50 backbone \cite{he2016deep} pretrained on ImageNet \cite{deng2009imagenet} is used as the feature extractor. For training, we employ the L1 loss across all methods, including our own. For the diffusion network, we adopt the same setup and strategy outlined in \cite{realgdsr}. This adjustment decreases smoothing and makes the filled \gls{gl:DSM} more coherent.

We employ the ADAM optimizer~\cite{kingma2014adam} with a base learning rate of $5 \times 10^{-5}$ and no weight decay. The batch size is set to 8 for training, using an NVIDIA A100 GPU. We stop training once the \gls{gl:RMSE} on the validation set has converged. Our model is implemented in PyTorch.

\subsection{Baselines}

\begin{itemize}
    \item Spline Interpolation: A traditional interpolation technique that estimates missing \glspl{gl:DSM} values by fitting spline functions to the available data points

    \item Diff-DEM\cite{Lo2024}: An adaptation of the Palette diffusion model \cite{sharia2022palette} for \gls{gl:DSM} void filling. The model uses a U-Net architecture with attention mechanisms in deeper layers. Inputs are processed as dual-channel $2 \times 128 \times 128$ images, incorporating the \gls{gl:DSM} containing voids and the step $t$ approximation of the target \gls{gl:DSM}.
    
    \item RSAGAN\cite{Yue2024}: A \gls{gl:GAN}-based framework, where generator architecture includes two encoders with gated convolutions \cite{GatedConvolutions2015} to extract features while considering void masks. A Pyramid, Cascading, and Deformable Convolution (PCD) alignment module \cite{PCDAlign} aligns image and \gls{gl:DEM} features. To address large-scale voids, RSAGAN employs the \gls{gl:ICAM} for preliminary void filling and the \gls{gl:TFG-RPAB} to refine the features by transferring image textures to topographic attributes.

\end{itemize}
\subsection{Evaluation Metrics} 

\label{sec:metrics}
We evaluate the models' performance by examining the \gls{gl:RMSE}, the \gls{gl:NMAD}, and the \gls{gl:MedAE}, which are all derived from per-pixel differences between prediction and ground truth.

\section{Results}

\newcolumntype{C}[1]{>{\centering\arraybackslash}m{#1}}

\begin{figure*}[!ht]
    \centering
    \begin{tabular}{C{30mm}C{30mm}C{30mm}C{30mm}C{30mm}}
        \toprule
        \textbf{Input} & \textbf{Guide} & \textbf{RSAGAN} & \textbf{Dfilled} & \textbf{Ground Truth} \\
        \midrule
        \includegraphics[height=3cm]{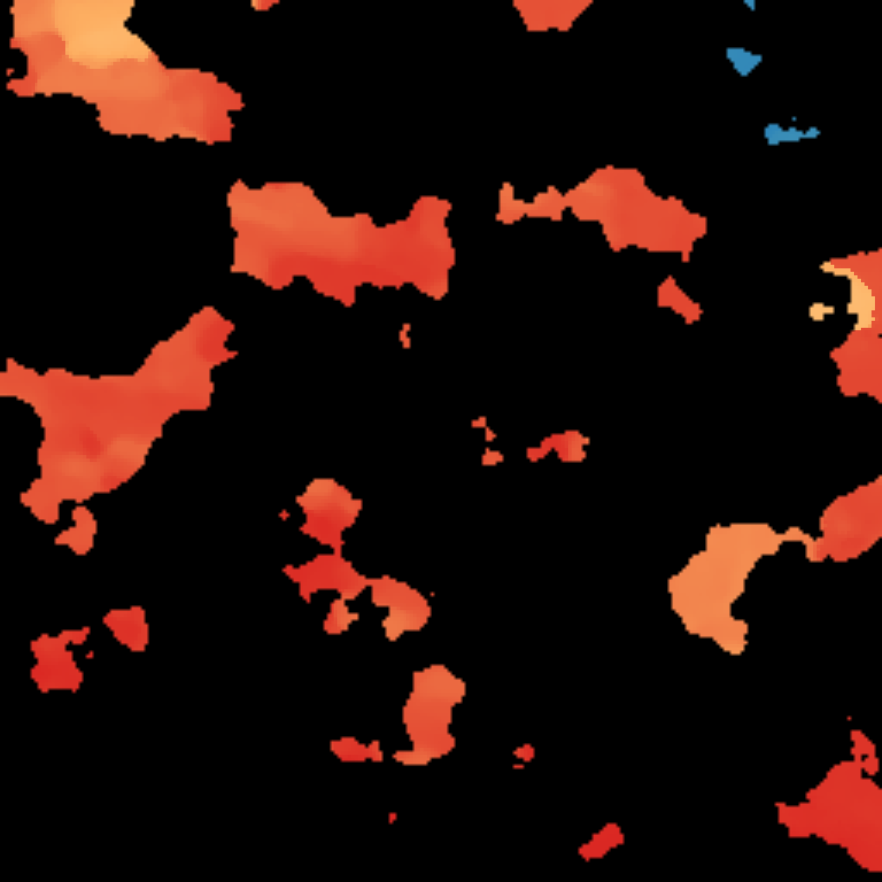} 
        & \includegraphics[height=3cm]{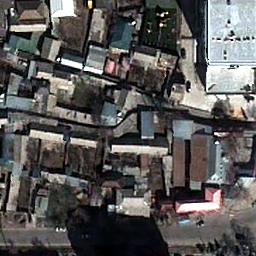} 
        & \includegraphics[height=3cm]{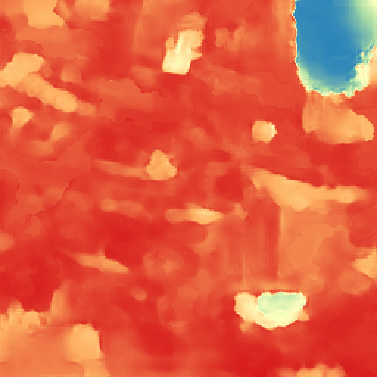} 
        & \includegraphics[height=3cm]{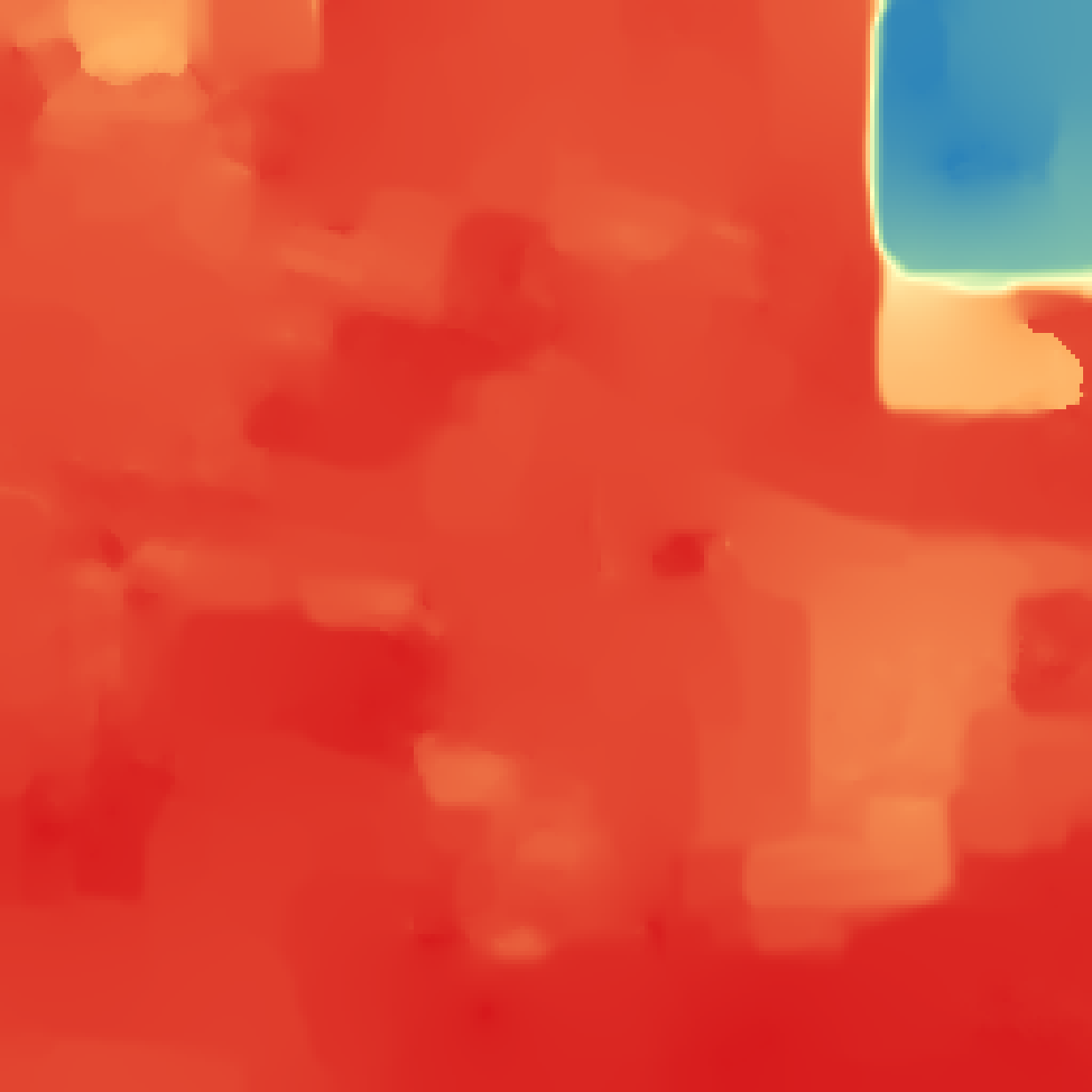} 
        & \includegraphics[height=3cm]{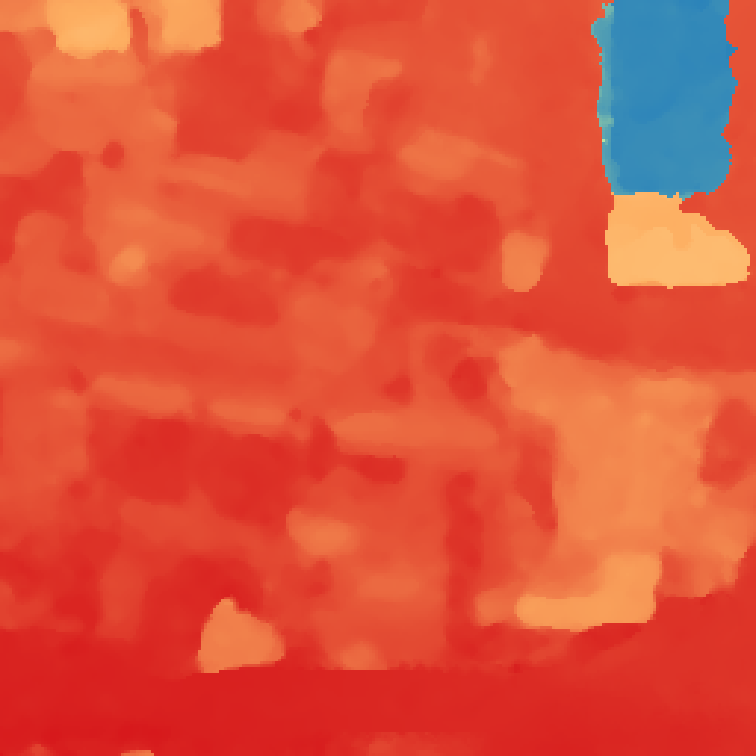} \\
        \bottomrule
    \end{tabular}
    \caption{Visual comparison of void filling results for large masks, where 60–80\% of the DSM area is void-filled. The Dfilled method effectively utilizes high-resolution guide images to reconstruct missing elevation data, outperforming RSAGAN in recovering complex terrain features and structural details.}
    \label{fig:large_masks}
\end{figure*}

\subsection{Comparisons with Prior Works}
Our proposed method, Dfilled, demonstrates superior performance in filling voids within \glspl{gl:DSM}, as evidenced by both quantitative and qualitative results. In \Cref{tab:comparison}, we compare four methods, Spline, Diff-DEM, RSAGAN, and Dfilled, across three different datasets (Real, Small Synthetic, and Large Synthetic), each trained on LaMa or Perlin masks. Dfilled obtains significantly lower RMSE, NMAD, and MedAE values compared to traditional interpolation (Spline) and state-of-the-art learning-based approaches (RSAGAN and Diff-DEM).

Interestingly, Diff-DEM, a recent single DSM inpainting approach, outperforms the guided RSAGAN on the Real dataset. However, Dfilled surpasses both methods across all three datasets. This underscores the robustness of our guided approach and its ability to leverage auxiliary information effectively during void filling. Although many methods benefit from being trained with our newly introduced Perlin mask, Dfilled achieves strong and consistent results irrespective of whether the LaMa or Perlin mask is used in training, illustrating that its performance does not heavily depend on the specific mask type and can readily adapt to varying void patterns.

Complementing these quantitative findings, \cref{fig:small_masks} and \cref{fig:large_masks} provide visual comparisons for voids of different sizes. In scenarios with smaller voids (see \cref{fig:small_masks}), Dfilled not only reconstructs the missing elevation data more accurately than RSAGAN but also produces smoother, more regularized outputs, minimizing artifacts and ensuring better continuity of terrain and structural details. For larger voids covering up to 60–80\% of the image (\cref{fig:large_masks}), Dfilled leverages high-resolution guide images to recover complex terrain and fine-scale features effectively. By contrast, RSAGAN frequently introduces artifacts and struggles to preserve fine details in both small and large void cases. Overall, these results highlight Dfilled’s adaptability, robustness, and ability to consistently produce realistic \gls{gl:DSM} reconstructions across a wide range of void sizes, making it a superior choice for demanding high-resolution applications.

\glsresetall

\subsection{Ablation Study}
\begin{table}[!ht]
    \begin{center}
    \caption{Contribution of each component in our network}
    \begin{tabular}{l|cc|c}
        \toprule
        & Refinement & Diffusion & RMSE [m] \\
        \midrule
        1 & & $\checkmark$ & 3.49 \\
        2 & $\checkmark$ & & 3.69 \\
        3 & $\checkmark$ & $\checkmark$ & \textbf{2.91} \\
        \bottomrule
    \end{tabular}
    \label{tab:ablation}
    \end{center}
\end{table}

\begin{figure}[!t]
    \centering
    \begin{subfigure}{.45\linewidth}
        \centering
        \begin{tikzpicture}
            \node[anchor=south west,inner sep=0] (img) at (0,0) {\includegraphics[width=\linewidth]{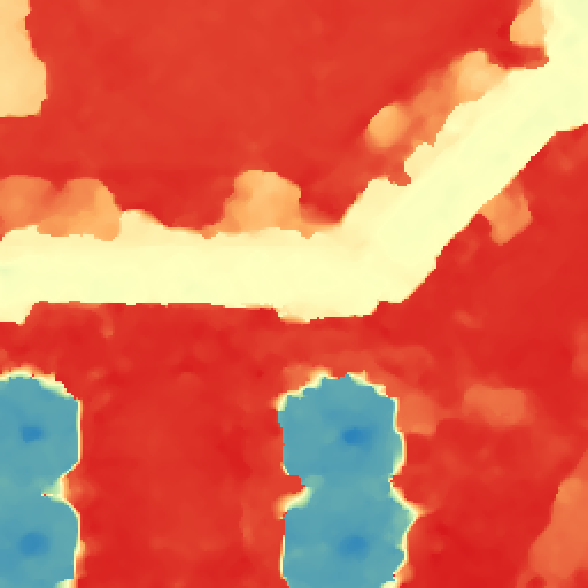}};
            \begin{scope}[x={(img.south east)}, y={(img.north west)}]
                \draw[green, thick] (0.75,0.02) rectangle (0.6,0.2);
            \end{scope}
        \end{tikzpicture}
        \caption{Only Diffusion}
        \label{tab:ablation_diffusion}
    \end{subfigure}
    \hfill
    \begin{subfigure}{.45\linewidth}
        \centering
        \begin{tikzpicture}
            \node[anchor=south west,inner sep=0] (img) at (0,0) {\includegraphics[width=\linewidth]{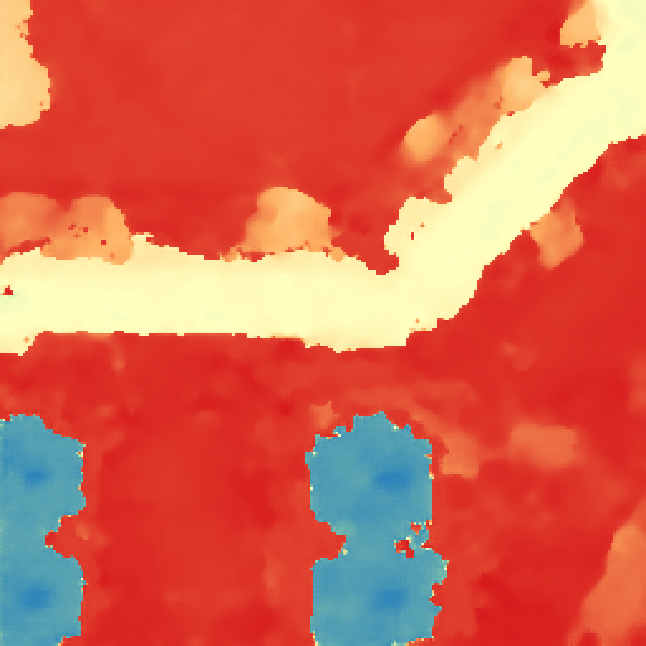}};
            \begin{scope}[x={(img.south east)}, y={(img.north west)}]
                \draw[green, thick] (0.75,0.02) rectangle (0.6,0.2);
            \end{scope}
        \end{tikzpicture}
        \caption{Only Refinement}
        \label{tab:ablation_refinement}
    \end{subfigure}
    \hfill
    \begin{subfigure}{.45\linewidth}
        \centering
        \begin{tikzpicture}
            \node[anchor=south west,inner sep=0] (img) at (0,0) {\includegraphics[width=\linewidth]{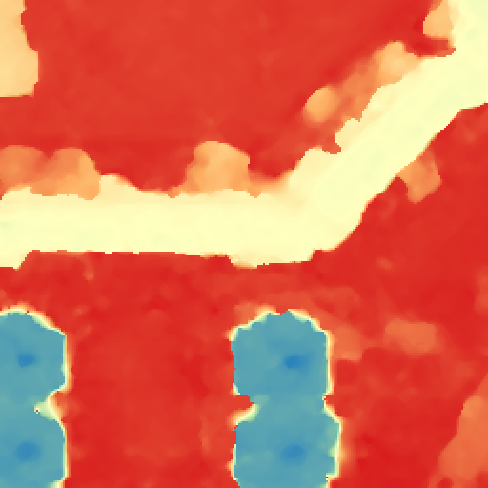}};
            \begin{scope}[x={(img.south east)}, y={(img.north west)}]
                \draw[green, thick] (0.75,0.02) rectangle (0.6,0.2);
            \end{scope}
        \end{tikzpicture}
        \caption{Refinement + Diffusion}
        \label{fig:line_profile}
    \end{subfigure}

    \caption{
        Visual comparison of ablation settings.
        (a) Using only Diffusion.
        (b) Using only Refinement.
        (c) Using both Refinement and Diffusion. Combining Refinement and Diffusion removes artifacts and smooths discontinuities.
    }
    \label{fig:ablation_study}
\end{figure}
To assess the contribution of each component in our model, we conducted an ablation study by systematically enabling the Refinement and Diffusion modules. As shown in \cref{tab:ablation}, using only the Diffusion module (row~1) achieves an RMSE of \SI{3.49}{\meter}, while using only the Refinement module (row~2) yields \SI{3.69}{\meter}. Enabling both modules (row~3) significantly reduces the RMSE to \SI{2.91}{\meter}, demonstrating the complementary nature of these components.

\Cref{fig:ablation_study} visually highlights the differences. Outputs with only the Refinement module (b) show noisy, dot-like artifacts, whereas adding the Diffusion module (c) removes these artifacts, producing smoother and more consistent results. This underscores the critical role of the Diffusion module in enhancing both accuracy and visual quality. Additionally, when combining both Refinement and Diffusion modules the resulted DSM is more regularized than the one produced without Refinement module (a).

\section{Conclusion}
In this paper, we introduced a novel approach for void filling in DSMs by adapting deep anisotropic diffusion models originally designed for super-resolution tasks. Our method redefines the problem using the heat equation and modifies the model to handle localized missing data through local refinement strategies for void initialization. This enables effective propagation of contextual information into voids while preserving critical structural details such as sharp building edges and smooth transitions in natural terrain.

To train our model on realistic missing data scenarios, we employed Perlin noise to generate inpainting masks that simulate natural void patterns commonly found in \glspl{gl:DSM}. Our method demonstrates robustness against large masks and various types of masks, making it effective across diverse landscapes, including complex urban environments. Extensive experiments on both simulated and real \gls{gl:DSM} datasets show that our approach outperforms traditional interpolation techniques and state-of-the-art deep learning methods. Specifically, our method excels in handling complex features and provides accurate, visually realistic results where conventional methods often fail. By utilizing edge-enhancing diffusion techniques, our proposed method enhances edge and structural information critical for maintaining terrain integrity in \glspl{gl:DSM}.

\section*{ACKNOWLEDGEMENTS}\label{ACKNOWLEDGEMENTS}
Special thanks are given to the GAF AG for the provision of the Pleiades 1B data.

{\small
\bibliographystyle{ieee_fullname}
\bibliography{egbib}
}

\end{document}